\documentclass[a4paper]{article}
\usepackage{INTERSPEECH2021}
\usepackage{cite}
\usepackage{balance}

\title{Optimizing Latency for Online Video Captioning\\Using Audio-Visual Transformers}
\name{Chiori Hori, Takaaki Hori, Jonathan Le Roux}
\address{
  Mitsubishi Electric Research Laboratories (MERL), Cambridge, MA, USA
}
\email{\{chori, thori, leroux\}@merl.com}

\begin{document}

\setlength{\abovedisplayskip}{7pt}
\setlength{\belowdisplayskip}{7pt}
\maketitle
\begin{abstract}
 Video captioning is an essential technology to understand scenes and describe events in natural language. To apply it to real-time monitoring, a system needs not only to describe events accurately but also to produce the captions as soon as possible. Low-latency captioning is needed to realize such functionality, but this research area for online video captioning has not been pursued yet. This paper proposes a novel approach to optimize each caption’s output timing based on a trade-off between latency and caption quality. An audio-visual Transformer is trained to generate ground-truth captions using only a small portion of all video frames, %
 and to mimic outputs of a pre-trained Transformer to which all the frames are given. A CNN-based timing detector is also trained to detect a proper output timing, where the captions generated by the two Transformers become sufficiently close to each other. With the jointly trained Transformer and timing detector, a caption can be generated in the early stages of an event-triggered video clip, as soon as an event happens or when it can be forecasted. Experiments with the ActivityNet Captions dataset show that our approach achieves 94\% of the caption quality of the upper bound given by the pre-trained Transformer using the entire video clips, using only 28\% of frames from the beginning. %
\end{abstract}

\section{Introduction}
At any time instant, countless events that happen in the real world are captured by cameras and stored as massive video data resources. To effectively retrieve such recordings, whether in offline or online settings, %
video captioning is an essential technology thanks to its ability to understand scenes and describe events in natural language.

Since the S2VT system~\cite{venugopalan2015sequence, venugopalan-etal-2015-translating} was first proposed, video captioning has been actively researched in the field of computer vision~\cite{yao2015describing, rohrbach2015long, pan2016jointly, yu2016video, otani2016learning} using sequence-to-sequence models in an end-to-end manner~\cite{bahdanau2014neural}. 
Its goal is to generate a video description (caption) about objects and events in a video clip.
To further leverage audio features to identify events, \cite{Hori_2017_ICCV} proposed the multimodal attention approach to fuse audio and visual features such as VGGish~\cite{hershey2017VGGish} and I3D~\cite{carreira2017quo} to generate video captions.
Such video clip captioning technologies have been expanded to offline video stream captioning technologies such as dense video captioning \cite{krishna2017dense} and progressive video description generator \cite{xiong2018move}, where all salient events in a video stream are temporally localized, and event-triggered captions are generated in a multi-thread manner. 
While all video captioning technologies had so far been based on LSTM, \cite{iashin2020abetter} successfully applied the Transformer~\cite{vaswani2017attention, karita2019comparative, dong2018speech} together with the audio-visual attention framework~\cite{Hori_2017_ICCV}. 
In that work, the audio-visual Transformer was tested using the ActivityNet Captions dataset~\cite{krishna2017dense} within an offline video captioning system and achieved the best performance for the dense video captioning task.
However, such offline video captioning technologies are not practical in real-time monitoring or surveillance systems, in which it is essential not only to describe events accurately but also to produce captions as soon as possible to find and report the events quickly. Low-latency captioning %
is required to realize such functionality, but this research area has not been pursued yet. 

This paper proposes a novel approach that optimizes the output timing for each caption based on a trade-off between latency and caption quality. We train a low-latency audio-visual Transformer %
composed of (1) a Transformer-based caption generator which tries to generate ground-truth captions after only seeing a small portion of all video frames, and also to mimic the outputs of a similar pre-trained caption generator that is allowed to see the entire video, and (2) a CNN-based timing detector that can find the best timing to output a caption, such that the captions ultimately generated by the above two Transformers become sufficiently close to each other. %

The proposed jointly-trained caption generator %
and timing detector can generate captions in an early stage of a video clip, as soon as an event happens. Additionally, this framework has the potential to forecast future events in online captions.
Furthermore, by combining multimodal sensing information, an event can be recognized at an earlier timing triggered by the earliest cue in one of the modalities without waiting for other cues in other modalities. For example, the proposed approach %
has the potential to generate captions earlier than a visual cue's timing based on the timing of an audio cue. Such a low-latency online video captioning using multimodal sensing information will contribute not only to retrieve events quickly but also to answer questions about scenes earlier \cite{hori_2019_ICASSP, Alamri_2019_CVPR}.

\section{Related work}
There are some works on low-latency end-to-end sentence generation for machine translation (MT) and automatic speech recognition (ASR). 
To realize real-time interpretation systems, simultaneous translation using greedy decoding was proposed and opened up the issue of streaming for neural MT (NMT)~\cite{cho2016can, gu-etal-2017-learning, ma-etal-2019-stacl, dalvi-etal-2018-incremental, arivazhagan-etal-2019-monotonic}; an emission point when a phrase is fully translated into a target language was incrementally determined. Another approach iteratively retranslates by concatenating subsequent words and updating the output~\cite{Niehues2018, arivazhagan-icassp2020-retrans}.
The goal is to generate a partial translation in the meantime before a full source sentence is translated. In contrast, our goal is to generate a full caption as soon as the system believes enough cues have been captured before seeing the entire video. 
Real-time ASR technology is also essential for applications such as closed captions. Some end-to-end systems regularize or penalize the emission delay using endpoint detection, and penalty terms that constrain alignments were proposed~\cite{li2020towards, sak2015fast, sainath2020emitting, yu2020fastemit}. 
There, the target is to generate a transcription slightly earlier from the end of an utterance.
In contrast, our target is to generate video captions as early as possible before the end of events.

In the field of computer vision, PickNet was proposed to find salient visual frames sufficient to generate video captions, where the target number of frames was given, and the captioning capability using only the selected frames was evaluated~\cite{chen2018less}. The paper mentions that it may be possible to apply PickNet to online captioning, showing a sample use case, but no quantitative evaluation was performed.
Another work relevant to online video captioning attempts to anticipate caption generation for future frames~\cite{hosseinzadeh2021video}. This approach exploited the current event features as a contextual feature and input them into a captioning module to generate future captions. This technology uses temporal dependency between events in a sequence.

\section{Online multi-modal captioning Transformer}
\subsection{Architecture}

\begin{figure}[t]
    \centering
    \includegraphics[width=7cm]{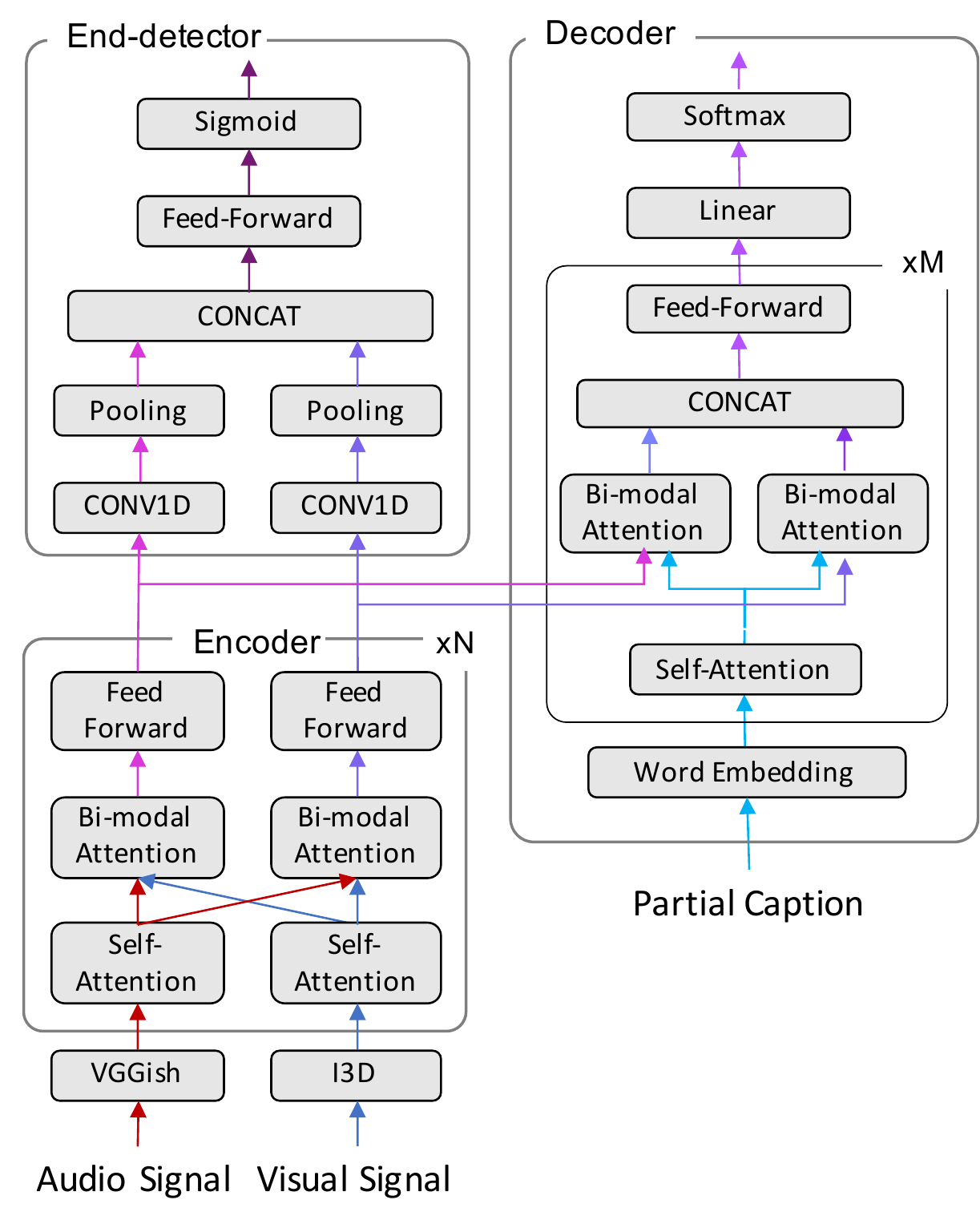}
    \vskip -2mm
    \caption{Online multi-modal captioning Transformer.}
    \label{fig:online-captioning-model}
    \vskip -3mm
\end{figure}

We describe the proposed low-latency video captioning model.
Figure \ref{fig:online-captioning-model} illustrates the model architecture, which consists of an audio-visual encoder, an end detector, and a caption decoder, where the encoder is shared by the detector and the decoder. Our model is based on the Transformer architecture~\cite{vaswani2017attention} and its multimodal extension~\cite{iashin2020abetter}, but it receives video and audio features in a streaming manner, and the end detector decides when to generate a caption for the feature sequence the model has received until that moment.

Given a video stream, the audio-visual encoder extracts VGGish and I3D features from the audio and video tracks, respectively, where the frame rate may be different on each track.
The sequences of audio and visual features from a starting point to the current time are fed to the encoder, and converted to hidden vector sequences through self-attention, bi-modal attention, and feed-forward layers. Typically, this encoder block is repeated $N$ times, e.g., $N=6$ or greater. The final encoded representation is obtained via the $N$-th encoder block.

Let $X^A$ and $X^V$ be audio and visual signals. First, the feature extraction module is applied to the input signals as
\begin{align}
A^0=\mathrm{VGGish}(X^A), ~~V^0=\mathrm{I3D}(X^V),
\end{align}
to obtain feature vector sequences corresponding to the VGGish and I3D features, respectively.
Each encoder block computes hidden vector sequences as
\begin{align}
    \bar{A}^n &= A^{n-1}+\mathrm{MHA}(A^{n-1}, A^{n-1}, A^{n-1}),\label{eq:enc_self1}\\
    \bar{V}^n &= V^{n-1}+\mathrm{MHA}(V^{n-1}, V^{n-1}, V^{n-1}),\label{eq:enc_self2}\\
    \tilde{A}^n &= \bar{A}^{n}+\mathrm{MHA}(\bar{A}^{n}, \bar{V}^{n}, \bar{V}^{n}),\label{eq:enc_bm1}\\
    \tilde{V}^n &= \bar{V}^{n}+\mathrm{MHA}(\bar{V}^{n}, \bar{A}^{n}, \bar{A}^{n}),\label{eq:enc_bm2}\\
    A^n&=\tilde{A}^{n}+\mathrm{FFN}(\tilde{A}^{n}), \\
    V^n&=\tilde{V}^{n}+\mathrm{FFN}(\tilde{V}^{n}),
\end{align}
where $\mathrm{MHA}$ and $\mathrm{FFN}$ denote multi-head attention and feed-forward network, respectively.
Layer normalization~\cite{ba2016layer} is applied before every $\mathrm{MHA}$ and $\mathrm{FFN}$ layers, but it is omitted from the equations for simplicity.
$\mathrm{MHA}$ takes three arguments, query, key, and value vector sequences~\cite{vaswani2017attention}. 
The self-attention layer extracts temporal dependency within each modality, where the arguments for $\mathrm{MHA}$ are all the same, i.e., $A^{n-1}$ or $V^{n-1}$, as in \eqref{eq:enc_self1} and \eqref{eq:enc_self2}.
The bi-modal attention layers further extract cross-modal dependency between audio and visual features, taking the keys and values from the other modality as in \eqref{eq:enc_bm1} and \eqref{eq:enc_bm2}.
After that, the feed-forward layers are applied in a point-wise manner.
The encoded representations for audio and visual features are obtained as $A^N$ and $V^N$.

The end detector receives the encoded representation based on the audio-visual information available at the moment. The role of the end detector is to decide whether the system should generate a caption or not for the given encoded features.
The detector first processes the encoded vector sequence from each modality with stacked 1D-convolution layers as
\begin{align}
    A_{c} = \mathrm{Conv1D}(A^N),~~V_{c} = \mathrm{Conv1D}(V^N).
\end{align}
Each time-convoluted sequences are then summarized into a single vector through pooling and concatenation operations:
\begin{align}
    H=\mathrm{Concat}(\mathrm{MeanPool}(A_c), \mathrm{MeanPool}(V_c))
\end{align}
A feed-forward layer $\mathrm{FFN}$ and sigmoid function $\sigma$ convert the summary vector to the probability of $d$, where $d$ indicates whether a relevant caption can be generated or not:
\begin{align}
    P(d=1|X^A,X^V)=\sigma(\mathrm{FFN}(H)).
\end{align}
Once the end detector provides a higher probability than a threshold, e.g., $P(d=1|X^A,X^V)>0.5$, the decoder generates a caption based on the encoded representation $(A^N,V^N)$.

The decoder iteratively predicts the next word from a starting token (\texttt{<sos>}).
At each iteration step, it receives a partial caption that has already been generated, and predicts the next word by applying $M$ decoder blocks and a prediction network, where each word is assumed to be converted to a word embedding vector. 

{\allowdisplaybreaks
Let $Y_i^0$ be partial caption $\texttt{<sos>},y_1,...,y_i$ after $i$ iterations.
Each decoder block has self-attention, bi-modal source attention, and feed-forward layers:
\begin{align}
    \bar{Y}_i^m &=Y_i^{m-1} + \mathrm{MHA}(Y_i^{m-1},Y_i^{m-1},Y_i^{m-1}),\label{eq:dec_self} \\
    \bar{Y}_i^{Am} &=\bar{Y}_i^m + \mathrm{MHA}(\bar{Y}_i^m, A^N, A^N), \label{eq:dec_src1}\\
    \bar{Y}_i^{Vm} &=\bar{Y}_i^m + \mathrm{MHA}(\bar{Y}_i^m, V^N, V^N), \label{eq:dec_src2}\\
    \tilde{Y}_i^m &=\mathrm{Concat}(\bar{Y}_i^{Am}, \bar{Y}_i^{Vm}), \label{eq:dec_ff1}\\
    Y_i^m &= \tilde{Y}_i^m + \mathrm{FFN}(\tilde{Y}_i^m). \label{eq:dec_ff2}
\end{align}
The self-attention layer converts the word vectors to high-level representations considering their temporal dependency in \eqref{eq:dec_self}.
The bi-modal source attention layers update the word representations based on the relevance to the encoded multi-modal representations in \eqref{eq:dec_src1} and \eqref{eq:dec_src2}.
A feed-forward layer is then applied to the outputs of the bi-modal attention layers in \eqref{eq:dec_ff1} and \eqref{eq:dec_ff2}.
Finally, a linear transform and a softmax operation are applied to the output of the $M$-th decoder block to obtain the probability distribution of the next word as
\begin{align}
    P(\mathbf{y}_{i+1}|Y_i,X^A,X^V)=\mathrm{Softmax}(\mathrm{Linear}(Y_i^M)), \\
    \hat{y}_{i+1} = \mathop{\mathrm{argmax}}_{y \in \mathcal{V}} P(\mathbf{y}_{i+1}=y|Y_i,X^A,X^V),
\end{align}
where $\mathcal{V}$ denotes the vocabulary.
}

After picking the one-best word $\hat{y}_{i+1}$, the partial caption is extended by adding the selected word to the previous partial caption as $Y_{i+1}=Y_i,\hat{y}_{i+1}$.
This is a greedy search process that ends if $\hat{y}_{i+1}=\texttt{<eos>}$, which represents an end token.
It is also possible to pick multiple words with highest probabilities and consider multiple candidates of captions according to the beam search technique.

Similar architectures have been used for dense video captioning tasks \cite{zhou2018end, iashin2020abetter}, where an event localization network is placed on top of the encoder similarly to our end detector.
A difference with those models is that the localization network is assumed to access all frames of the video and chooses a set of regions, which potentially includes specific events, while our end detector can access only partial frames from the beginning or a certain point to the current frame and detect a timing at which the system should emit the caption.
Thus, our model is designed and trained for online captioning.

\subsection{Training}
\label{sec:training}
We learn the multi-modal encoder, the end detector, and the caption decoder jointly, so that the model achieves a caption quality comparable to that for a complete video, even if the given video is shorter than the original one by truncating the later part. 

\begin{figure}[t]
    \centering
    \includegraphics[width=8 cm]{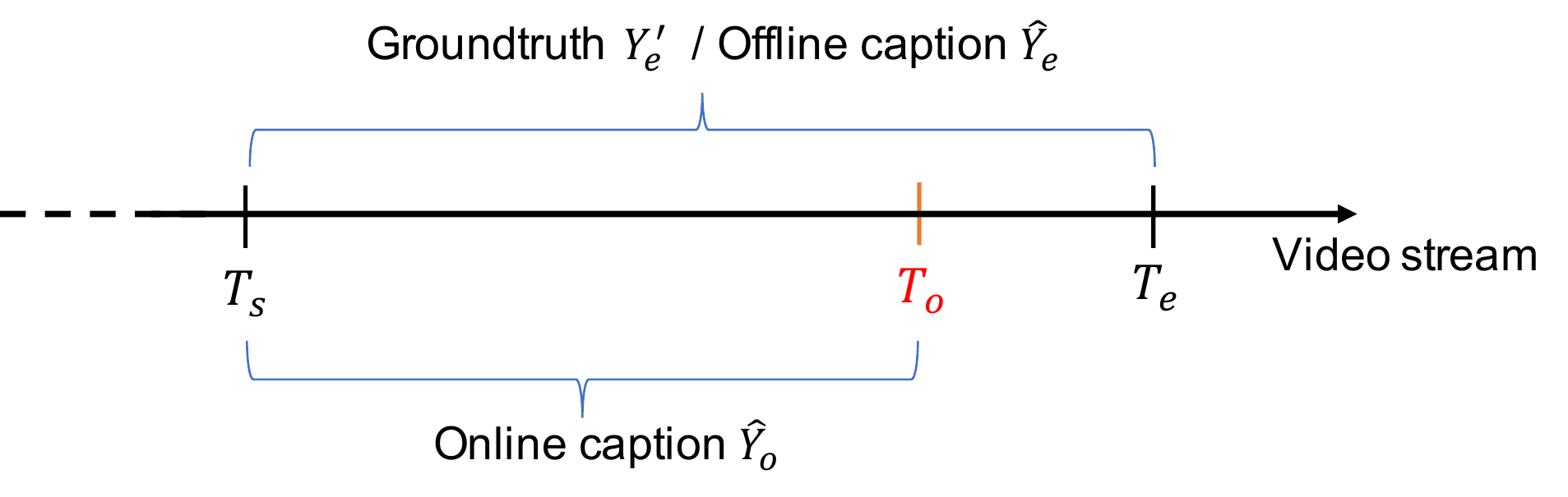}
    \vskip -1mm
    \caption{Online and offline captions.}
    \label{fig:captioning-region}
    \vskip -3mm
\end{figure}

Two types of loss functions are combined, a captioning loss to improve the caption quality and an end detection loss to detect a right timing to emit a caption.
Figure \ref{fig:captioning-region} shows an example of video stream, where an event has started at time $T_s$ and ends at $T_e$, and is associated with ground-truth caption $Y_e'$.
If time $T_o$ is picked as the emission timing, the captioning decoder generates a caption based on the multi-modal input signal $X_{T_s:T_o}=(X^A_{T_s:T_o}, X^V_{T_s:T_o})$.

The captioning loss is based on a standard cross entropy loss for the ground-truth caption $Y_e'$,
\begin{align}
    \mathcal{L}_{CE}=-\log P(Y_e'|X_{T_s:T_o}; \theta_C),
\end{align}
and a Kullback–Leibler (KL) divergence loss between predictions from a pre-trained model allowed to process the complete video and the target model that can only process incomplete videos, i.e.,
\begin{align}
    \mathcal{L}_{KL}=-\sum_{i=1}^{|Y_e'|}\sum_{y\in \mathcal{V}}P&(y|Y_{e,i}',X_{T_s:T_e};\bar{\theta}_C)\nonumber \\
    & \log P(y|Y_{e,i}',X_{T_s:T_o};\theta_C).
\end{align}
This is a student-teacher learning approach to exploit another model's superior description power~\cite{hori2019student}, where the teacher model $\bar{\theta}_C$ predicts a caption using entire video clip $X_{T_s:T_e}$ and the student model $\theta_C$ tries to mimic the teacher's predictions using only the truncated video clip $X_{T_s:T_o}$. This makes the training more stable and achieves better performance.

The end detection loss is based on a binary cross entropy for appropriate timings. In general, however, such timing information does not exist in the training data set.
In this work, we decide the right timing based on whether or not the captioning decoder can generate a relevant caption, that is, a caption sufficiently close to the ground-truth $Y_e'$ or the caption $\hat{Y}_e$ generated for the entire video clip $X_{T_s:T_e}$ using the pre-trained model $\bar{\theta}_C$.
The detection loss is computed as
\begin{align}
    \mathcal{L}_D=-\log P(d|X_{T_s:T_o};\theta_D),
\end{align}
where $d$ is determined based on
\begin{align}
    d=%
    \begin{cases}
         1 & \mathrm{if}~\max(\mathrm{Sim}(Y_e',\hat{Y}_o), \mathrm{Sim}(\hat{Y}_e,\hat{Y}_o)) \geq S,\\
         0 & \mathrm{otherwise},
    \end{cases}
    \label{eq:detection}
\end{align}
where $\mathrm{Sim}(\cdot,\cdot)$ denotes a similarity measure between two word sequences. In this work, we use word accuracy computed in a teacher-forcing manner. $S \in (0, 1]$ is a pre-determined threshold which judges whether or not the online caption $\hat{Y}_o$ is sufficiently close to the references $Y_e'$ and $\hat{Y}_e$.

The training process for model $\theta=(\theta_C,\theta_D)$ repeats the following steps:
\begin{enumerate}
    \item Sample $T_o \sim \mathrm{Uniform}(T_s,T_e)$,
    \item Compute loss $
        \mathcal{L}=\alpha \mathcal{L}_{CE} + \beta\mathcal{L}_{KL} + \gamma \mathcal{L}_D$,
    \item  Update $\theta$ using $\nabla_\theta\mathcal{L}$.
\end{enumerate}

\subsection{Inference}
The inference is performed in two steps:
\begin{enumerate}
    \item Find $\hat{T}_o$ that first satisfies $P(d=1|X_{T_s:\hat{T}_o};\theta_D)>F$,
    \item Generate a caption based on
    \begin{align}
       \hspace{-2cm} \hat{Y}_o=\mathop{\mathrm{argmax}}_{Y\in \mathcal{V}^*}P(Y|X_{T_s:\hat{T}_o};\theta_C),
    \end{align}
\end{enumerate}
where $F$ is a pre-determined threshold to control the sensitivity of end detection. Note that we assume that $T_s$ is already determined.

\begin{figure*}[t]
    \vskip -2mm
    \centering
    \includegraphics[width=13cm]{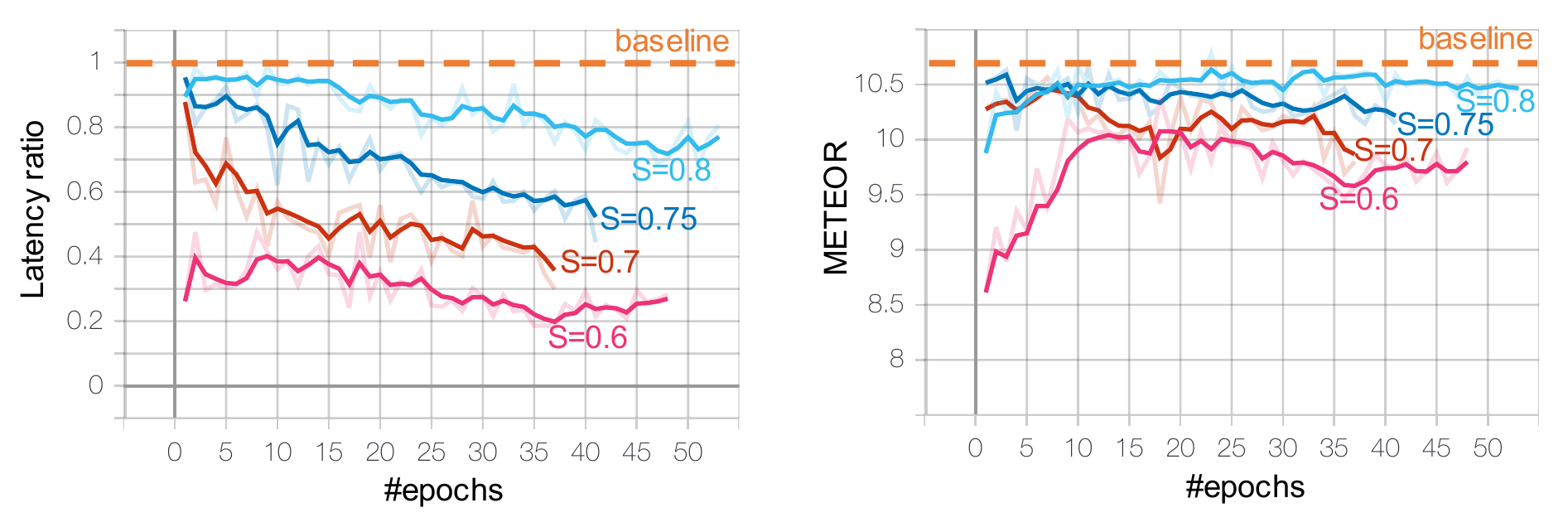}
    \vskip -3mm
    \caption{Latency ratio (left) and METEOR score (right) in training with different thresholds $S$ in \eqref{eq:detection}.}
    \label{fig:learning-curves}
    \vskip -3mm
\end{figure*}

\section{Experiments}
We evaluate our low-latency caption generation method using the ActivityNet Captions dataset~\cite{krishna2017dense}, which consists of 100k caption sentences associated with temporal localization information based on 20k YouTube videos.
Although the conventional video description dataset MSVD (YouTube2Text)~\cite{guadarrama2013youtube2text} and MSR-VTT~\cite{MSR-VTT:CVPR16} have 41 and 20 ground-truth captions for each video clip respectively, ActivityNet only has one for each event.
The dataset is split into 50\%, 25\%, and 25\% for training, validation, and testing. However, since the ground-truth captions for the test set are not available, we split the validation set into two subsets on which we report the performance as done in a prior study \cite{iashin2020abetter}.
The average duration of a video clip is 35.5, 37.7, and 40.2 seconds for the training set and the validation subsets 1 and 2, respectively.
We used VGGish and I3D features provided by the author of \cite{iashin2020abetter}.
The VGGish features were configured to form a 128-dimensional vector sequence for the audio track of each video, where each audio frame corresponds to a 0.96 s segment without overlap. The I3D features were configured to form a 1024-dimensional vector sequence for the video track, where each visual frame corresponds to a 2.56 s segment without overlap. 

A multi-modal Transformer was first trained with entire video clips and their ground-truth captions. This model was used as a baseline and teacher model.
We used $N=2$ encoder blocks and $M=2$ decoder blocks, and the number of attention heads was 4.
The vocabulary size was 10,172, and the dimension of word embedding vectors was 300.

The proposed model for online captioning was trained with incomplete video clips according to the steps in Section \ref{sec:training}. The architecture was the same as the baseline/teacher model except for the addition of the end detector.
In the training process, we consistently used $\alpha\!=\!\beta\!=\!\gamma\!=\!1/3$ for the loss function.
The dimensions of hidden activations in audio and visual attention layers were 128 and 1024, respectively.
The dropout rate was set to 0.1, and a label smoothing technique was also applied.
The end detector had 2 stacked 1D-convolution layers, with a ReLU non-linearity in between.
The performance was measured by BLEU3, BLEU4, and METEOR scores.

Figure \ref{fig:learning-curves} shows the latency ratio (left) and METEOR scores (right) on validation subset 1 when training with different $S$ in \eqref{eq:detection}.
The latency ratio indicates the ratio of the video duration used for captioning to the duration of the original video clip. With the baseline model, the latency ratio is always 1, which means all frames are used to generate captions.
With our proposed method, the latency ratio and METEOR scores change depending on the value of $S$, where a larger $S$ gives a stricter condition on the caption accuracy, resulting in later detection, while a smaller $S$ results in earlier detection.
As learning proceeds, the latency ratio gradually decreases, but the METEOR score tends to maintain high values close to the baseline.
This result demonstrates that the learning process works to reduce the latency while maintaining caption quality.

Table \ref{table:performance} compares captioning methods in BLEU and METEOR scores on validation subset 1.
The model selected for evaluation was trained with $S=0.6$ and had the best METEOR score on validation subset 2. We controlled the latency with the detection threshold $F$. As shown in the table, our proposed method at a 55\% latency achieves 10.45 METEOR score with only a small degradation, which corresponds to 98\% of the baseline score $10.67$ \cite{iashin2020abetter}. 
It also achieves $10.00$ METEOR score at a 28\% latency, which corresponds to 94\% of the baseline.
We also evaluated a naive method which takes video frames from the beginning with a fixed ratio to the original video length and runs the baseline captioning on the truncated video clip. The results show that the proposed approach clearly outperforms the naive method at an equivalent latency. 

The table also includes the results for a unimodal Transformer that receives only the visual feature. The results show that the proposed method works for the visual feature only, but the performance is degraded due to the lack of the audio feature. This result indicates that the audio feature is essential even in the proposed low-latency method. 

\begin{table}[t]
\centering
\caption{Performance of baseline and proposed systems. The scores are averaged on the captions in validation subset 1. The average duration of a video clip is 37.7 seconds. ST denotes Student-Teacher learning. }
\label{table:performance}
\vskip -3mm
\resizebox{.99\linewidth}{!}{
\begin{tabular}{lccccc}
\toprule
            Method & Latency  & BLEU-3 & BLEU-4 & METEOR \\
\cmidrule(lr){1-1}\cmidrule(lr){2-2}\cmidrule(lr){3-5}
Baseline \cite{iashin2020abetter} &    100\%   & 4.66 & 2.05 & 10.67  \\
\midrule
Naive method           &  \phantom{1}55\%  &  4.31 & 1.76 & 10.20  \\
Naive method           &  \phantom{1}33\%  &  3.69 & 1.37 & 9.59  \\
\midrule
Proposed (w/o ST)      &  \phantom{1}55\%  &  4.22 &  1.77  &  10.38  \\
Proposed               &  \phantom{1}56\%  &  {\bf 4.40}  &  {\bf 1.82} & {\bf 10.45} \\
Proposed (w/o ST)      &  \phantom{1}29\%  &  3.75     &     1.52    &    9.93     \\
Proposed               &  \phantom{1}28\%  & {\bf 3.84}  &  {\bf 1.57} & {\bf 10.00} \\
\midrule
Baseline (visual only) &          100\%    & 4.08 & 1.80 & 10.21 \\
Proposed (visual only) & \phantom{1}54\%   & 3.82 & 1.61 & 10.05 \\
Proposed (visual only) & \phantom{1}30\%   & 3.45 & 1.42 &  9.71 \\
\bottomrule
\end{tabular}
}
\vskip -4mm
\end{table}

\vskip -5mm
\section{Conclusions}
 In this paper, we proposed a low-latency audio-visual captioning method, which describes events accurately and quickly without waiting for the end of video clips.
 The proposed method optimizes each caption’s output timing based on a trade-off between latency and caption quality. 
 We have demonstrated that the proposed system can generate captions in early stages of  event-triggered video clips, achieving 94\% of the caption quality of the upper bound given by a Transformer processing the entire video clips, using only 28\% of frames (10.6 seconds) on average from the beginning.
\balance 
\bibliographystyle{IEEEtran}
\bibliography{mybib}

\end{document}